\pdfoutput=1

\documentclass[11pt]{article}

\usepackage[preprint]{acl}

\usepackage{times}
\usepackage{latexsym}
\usepackage{subfigure}
\usepackage{amsmath}
\usepackage{amsfonts}
\usepackage{enumitem}
\usepackage{multirow}
\usepackage{colortbl} 
\usepackage{xcolor}   
\usepackage{booktabs}
\usepackage{adjustbox}
\usepackage{makecell}
\usepackage{hyperref}
\usepackage[linesnumbered,ruled,vlined]{algorithm2e}
\usepackage[T1]{fontenc}

\usepackage[utf8]{inputenc}

\usepackage{microtype}

\usepackage{inconsolata}

\usepackage{graphicx}

\newcommand\ourmethod{\texttt{LANCE}\xspace}

\title{Language Models as Continuous Self-Evolving Data Engineers}

\author{
 Peidong~Wang$^1$, Ming~Wang$^1$, Zhiming~Ma$^2$, \\ \textbf{Xiaocui~Yang$^1$, Shi~Feng$^{1\dag}$, Daling~Wang$^1$, Yifei~Zhang$^1$, Kaisong~Song$^{1,3}$}  \hspace{0.4em} \\
 $^1$School of Computer Science and Engineering, Northeastern University, Shenyang, China \\
 $^2$China Mobile Internet Company Limited, Guangzhou, China \\
 $^3$Alibaba Group, Hangzhou, China \\
\texttt{pdongwang@163.com} \quad
\texttt{sci.m.wang@gmail.com} \\
\texttt{mazhiming312@outlook.com} \quad
\texttt{kaisong.sks@alibaba-inc.com}\\
\texttt{\{yangxiaocui, fengshi, wangdaling, zhangyifei\}@cse.neu.edu.cn}\\
}

\begin{document}
\maketitle
\begin{abstract}
Large Language Models (LLMs) have demonstrated remarkable capabilities on various tasks, while the further evolvement is limited to the lack of high-quality training data. In addition, traditional training approaches rely too much on expert-labeled data, setting a ceiling on the performance of LLMs. To address this issue, we propose a novel paradigm named \ourmethod (\textbf{LAN}guage models as \textbf{C}ontinuous self-\textbf{E}volving data engineers) that enables LLMs to train themselves by autonomously generating, cleaning, reviewing, and annotating data with preference information. Our approach demonstrates that LLMs can serve as continuous self-evolving data engineers, significantly reducing the time and cost of the post-training data construction. Through iterative fine-tuning on Qwen2 series models, we validate the effectiveness of \ourmethod across various tasks, showing that it can maintain high-quality data generation and continuously improve model performance. Across multiple benchmark dimensions, \ourmethod results in an average score enhancement of \textbf{3.64} for Qwen2-7B and \textbf{1.75} for Qwen2-7B-Instruct. This training paradigm with autonomous data construction not only reduces the reliance on human experts or external models but also ensures that the data aligns with human preferences, paving the way for the development of future superintelligent systems that can exceed human capabilities. Codes are available at: \url{https://github.com/Control-derek/LANCE}.
\end{abstract}

\def\thefootnote{\dag}
\footnotetext[1]{Corresponding author.}

\def\thefootnote{\arabic{footnote}}

\section{Introduction}
Large Language Models (LLMs) have exhibited extraordinary proficiency in tackling diverse tasks, such as natural language understanding, logical reasoning, code generation, and mathematical reasoning \citep{dubey2024llama, achiam2023gpt, liu2024deepseek}. These advancements are largely attributed to instruction tuning \citep{wei2022finetuned} and Reinforcement Learning from Human Feedback (RLHF) \citep{stiennon2020learning}, which have significantly improved the performance of LLMs using quality data. High-quality data not only enhances the precision and reliability of models but also ensures that the outputs are more aligned with human values and preferences \citep{wang2023data}. 

\begin{figure}[t]
    \centering
    \includegraphics[width=\linewidth]{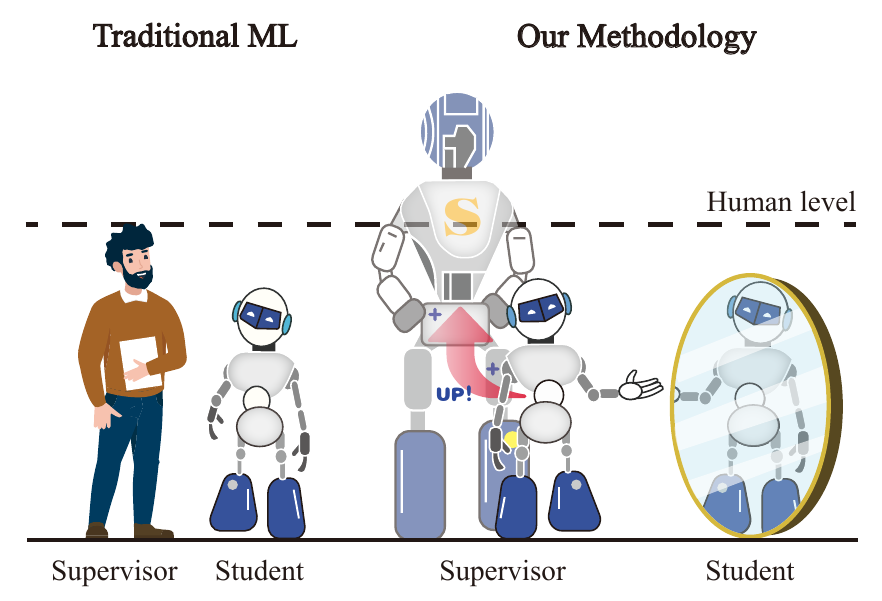}
    \caption{\textbf{An illustration of our methodology.} Traditional ML focuses on the setting where humans supervise models that are weaker than humans. Our methodology explores the scenario where models self-supervise, which may be a reliable path to superintelligence.}
    \label{robot}
    \vspace{-1em}
\end{figure}

However, with the rapid development of LLMs, the acquisition of high-quality data becomes increasingly challenging \citep{penedo2023the}. On one hand, the expansion of LLM training data has likely exhausted previously available high-quality sources \citep{villalobos2024position}, creating an ongoing demand for new data to enhance model performance. There is a constant need for new high-quality data to enhance LLMs' intelligence. On the other hand, relying on human experts to get high-quality data is time-consuming and costly, limiting efficiencies and making it challenging to keep up with the rapid demands of LLMs \citep{villalobos2024position}. Additionally, for future superintelligence that surpasses human capabilities \citep{burns2024weaktostrong}, human supervisory signals may be of limited help, and the quality of human-generated data may be one of the bottlenecks in the emergence of superintelligence \citep{burns2024weaktostrong}. Some research \citep{lee2024llm2llm, dai2025auggpt, chen2024alpagasus} suggests leveraging synthetic data from teacher LLMs to train student LLMs, yet this approach hinges on supervised signals from external models. Conversely, other studies \citep{zelikman2022star, wang-etal-2023-self-instruct, gulcehre2023reinforced} advocate for the use of data constructed by LLM to iteratively fine-tune the model itself. However, these methods do not comprehensively address the entire data construction lifecycle for the post-training \citep{dubey2024llama} of LLMs.

To address these challenges, we propose \ourmethod, a novel training paradigm empowering LLMs to autonomously generate, clean, review, and annotate data with preference information for self-training. This approach exemplifies how language models can operate as continuous self-evolving data engineers, minimizing human intervention and external resources. The process begins with the model reviewing the seed dataset. For lower-quality data, it generates new instructions and responses to explore similar themes or forms to construct instruction data, compensating for distributional deficiencies in the seed dataset. For higher-quality data, the model generates new responses that appear correct but are expressively flawed and contain misleading information, forming preference data to enhance model response quality and reduce hallucinations. The model reviews the newly generated data after calling the data filtering tool for initial cleaning to ensure high-quality instruction data and accurate preference pairs, using it for self-training. This iterative process can be repeated to continuously improve the model's performance. \ourmethod autonomously constructs data for model post-training, eliminating the need for human involvement or external models, markedly reducing time and cost compared to traditional methods. An illustration of our methodology is provided in Figure \ref{robot}. This self-supervised approach represents a paradigm shift from traditional human-supervised learning, where models are constrained by human expertise, to autonomous self-evolution, potentially paving the way for superintelligence development.

To evaluate our method, we conduct iterative fine-tuning on different models and assess their capabilities in various aspects such as scientific reasoning, commonsense reasoning, knowledge understanding, complex problem-solving, and mathematical reasoning. We find that even with iterative processing on a small dataset, the average performance of the model across various tasks continues to rise, and individual evaluation metrics either remain stable or show improvement. The experimental results demonstrate \ourmethod's effectiveness in enabling continuous model improvement through autonomous data processing. The consistent performance gains across multiple reasoning tasks suggest the potential of this self-evolving approach for advancing model intelligence. Importantly, our approach ensures efficient and cost-effective data generation, markedly reducing the reliance on human intervention or external models. This autonomous capability represents a crucial step toward developing systems that can potentially surpass human-level performance, contributing valuable insights to the ongoing research in superintelligence.

In summary, our contributions are as follows: (1) We propose \textbf{\ourmethod}, \textbf{a novel training paradigm}, which enables LLMs to autonomously generate, clean, review, and annotate data for self-improvement, substantially reducing time and expense in post-training data preparation. (2) \ourmethod autonomously manages the entire post-training data construction process, enhancing data generation efficiency and quality while \textbf{boosting model performance across diverse tasks}. (3) \ourmethod \textbf{enhances mathematical reasoning}, improving both elementary and advanced tasks as well as multilingual proficiency, despite relying solely on general-purpose data for training data generation.

\section{Related Work}

\begin{figure*}[th]
    \centering
    \includegraphics[width=1.0\linewidth]{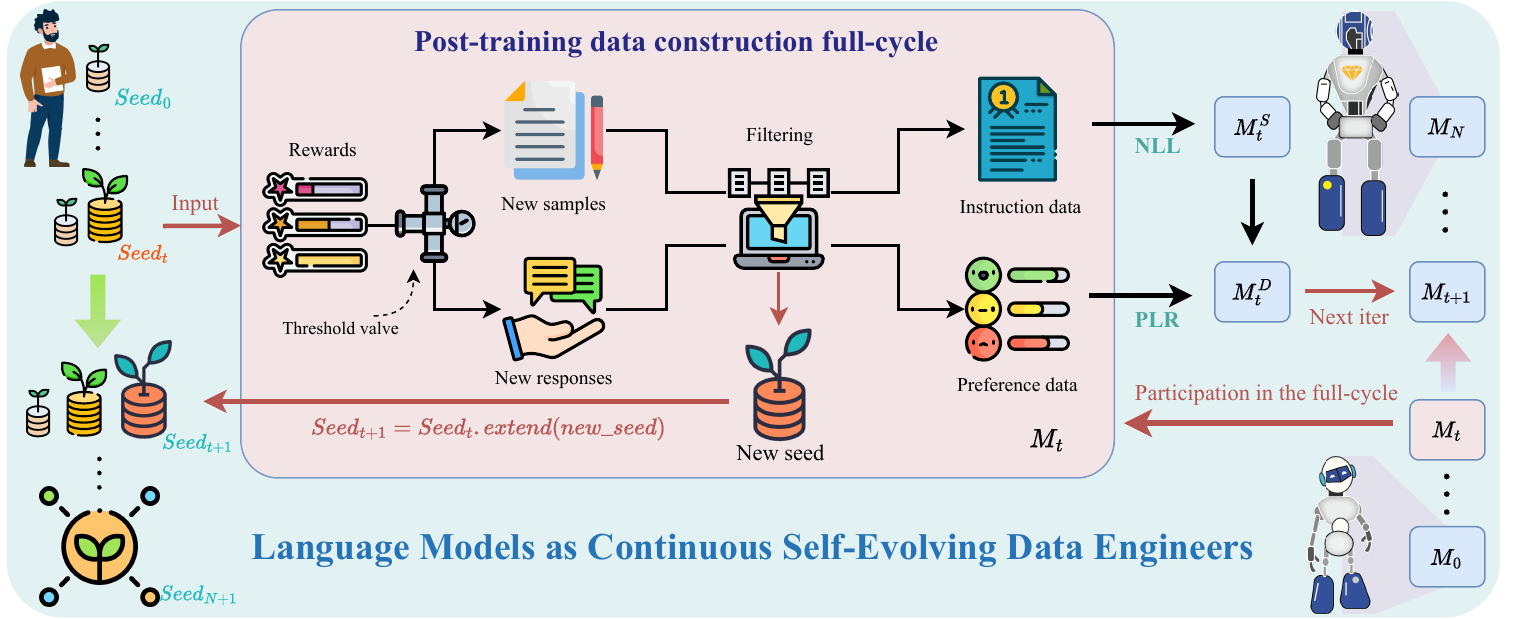}
    \caption{\textbf{Overview of \ourmethod.} The cycle begins at $t=0$ with pre-annotated seed dataset $Seed_{0}$. At each time step $t$, model $M_t$ generates new instruction and preference data from $Seed_{t}$ via \textbf{Post-training data construction full-cycle}. $M_t$ is fine-tuned on instruction data (NLL) to create $M_t^S$, then on preference data (PLR) to produce $M_t^D$. In the next iteration, $M_t^D$ becomes $M_{t+1}$, and new samples are merged into $Seed_{t}$ to form $Seed_{t+1}$.}
    \label{fig:overview}
\end{figure*}

\subsection{Post-Training}
After the pre-training, LLMs enter a post-training stage for instruction-following, alignment with human preferences, and enhancing specific skills (e.g., coding and reasoning) \citep{dubey2024llama}. Common methods include Supervised Fine-Tuning (SFT) \citep{wei2022finetuned} and preference learning \citep{stiennon2020learning, rafailov2024direct}. However, these approaches depend heavily on the availability and quality of human-annotated data.

\subsection{Data Augmentation for LLMs}
Data augmentation techniques enhance LLMs training data. For example, Alpaca \citep{alpaca} uses text-davinci-003 \citep{ouyang2022training} to generate data for fine-tuning a 7B LLaMA \citep{touvron2023llama} model. CoAnnotating \citep{li2023coannotating} employs human-LLM collaboration for efficient annotation. ReST-MCTS* \citep{zhang2024rest} integrates process reward with MCTS* to collect higher-quality reasoning traces. However, this approach is limited to tasks requiring ground-truth data. These methods often rely on external models, human intervention, or specific task constraints, increasing costs and limiting scalability.

\subsection{Self-Evolving LLMs}
Self-instruct \citep{wang-etal-2023-self-instruct} harnesses the LLM itself to construct instruction data, iteratively enhancing the model's instruction-following capability through SFT. Instruction backtranslation \citep{li2024selfalignment} augments unlabeled data with an instruction prediction model but requires substantial input. 
Self-rewarding \citep{yuan2024selfrewarding} employs the LLM itself as a reward model, iteratively training with DPO \citep{rafailov2024direct}, improving both the policy and reward models.
SPIN \citep{chen2024selfplay} aligns models with human preferences through self-play but is limited by data distribution. Our approach autonomously constructs a targeted dataset from seed data. By leveraging the LLM itself as a dynamic reward model, it addresses seed data deficiencies, enabling efficient data generation and model optimization. This process supports self-evolution without large-scale given data, expert-annotation, or external reward models.

\section{Methodology}

Figure \ref{fig:overview} illustrates an overview of our approach. Starting with a small seed dataset, the model generates instruction data and preference data through the pipeline, which is used to optimize the model via Negative Log-Likelihood (NLL) and Preference-driven Likelihood Ratio (PLR). The resulting model serves as the starting point for the next iteration. By repeating this cycle, the model’s overall performance is continuously improved.

\subsection{Preliminaries}

\subsubsection{Supervised Fine-Tuning}
Supervised Fine-Tuning (SFT), also known as instruction tuning, is typically applied to a pre-trained LLM to enhance its ability to understand and follow instructions, improving its performance on specific tasks. For a given model $M_t$, the training dataset $D^S=\{(x_i, y_i)\}_{i=1}^N$ consists of N instruction-response pairs $(x, y)$. During SFT, the LLM is trained to minimize the NLL loss:

\vspace{-2ex}
\begin{equation}
L_{S} = -\mathbb{E}_{D^S}[\sum_{k=1}^{T}log(M_t(y_{k}|y_{<k},x))]
\label{eq:sft}
\end{equation}

\noindent where $T$ represents the length of response $y$. From Equation \ref{eq:sft}, $L_{S}$ attains its minimum when the distribution of $M_t$ coincides with the conditional probability distribution $p(y|x)$ of the responses in $D^S$.

\subsubsection{Reinforcement Learning from Human Feedback}
Reinforcement Learning from Human Feedback (RLHF) trains models to learn human preferences from human feedback through reinforcement learning, enabling responses that align more closely with human expectations. This approach plays a crucial role in building safe, high-performance, and controllable AI systems \citep{ouyang2022training}.

Direct Preference Optimization (DPO) \citep{rafailov2024direct} eliminates the need for a reward model by directly using human preferences. Given the SFT model \( M_t^S \) and the model under optimization \( M_t^{\theta} \), the preference dataset \( D^P = \{ (x_i, y_i^w, y_i^l)\}_{i=1}^N  \) is used for training, where \( y^w \) and \( y^l \) denote the preferred and dispreferred responses, aiming to minimize the PLR loss:

\vspace{-2ex}
\begin{equation}
\hspace*{-0.3em}
{\text{$L_{D} = - \mathbb{E}_{D^P}[log~\sigma(\hat{r}(x,y^w) - \hat{r}(x,y^l))]$}}
\label{eq:dpo}
\end{equation}

In this context, $\hat{r}(x,y^w)$ and $\hat{r}(x,y^l)$ represent the reward values assigned by the model under optimization, \( M_t^{\theta} \), and the reference model, \( M_t^{S} \), to the preferred response $y^w$ and the dispreferred response $y^l$, respectively. They are computed as:

\vspace{-1ex}
\begin{equation}
\hat{r}(x,y) = \beta log\frac{M_t^{\theta}(y|x)}{M_t^S(y|x)}
\label{eq:reward}
\end{equation}

\noindent where $\beta > 0$ controls the deviation from the base reference policy, namely the initial SFT model $M_t^{S}$. 

\subsection{Initialization}

\textbf{Seed Data} Our seed data consists of two components: (i) A small labeled dataset, which serves as the initial foundation for the model to sample new data. (ii) A review dataset with prompts containing an instruction and a response, and outputs detailing the rationale and score. This setup enhances the model's proficiency in reviewing data, ensuring smooth iteration progression. The two components are randomly mixed to form the seed data.

\noindent \textbf{Large Language Model} In the iterative loop, we use an LLM that has undergone SFT on the seed data as our initial model \(M_0\). This SFT process equips the model with a foundational level of instruction-following and review capabilities. The model, starting from \(M_0\), continues to improve through subsequent iterations, evolving into \(M_1\) $\to$ $\cdots$ $\to$ \(M_t\) $\to$ $\cdots$ $\to$ \(M_N\), where \(t\) denotes the current iteration, with the process beginning at \(t = 0\) and concluding at \(t = N\).

\subsection{Post-training data construction} \label{sec:data_construction}

First, the seed data is reviewed using a language model to assess quality and identify deficiencies. Next, reward-based generation is employed to either enhance low-quality data or create preference pairs for high-quality data. Finally, the generated data undergoes rigorous filtering and is selectively added to the training datasets, ensuring high-quality instruction tuning and preference learning.

\subsubsection{Review Seed data}
The distribution of examples in an instruction tuning dataset is often uneven, with some topics or tasks having lower-quality instruction data than others. When such data is used for training, it may result in a language model with potential deficiencies in certain capabilities. \citet{zheng2023judging} demonstrated the potential of LLM as a judge, showing that a well-trained LLM can achieve high agreement with human evaluations. Based on these insights, we utilize the language model $M_t$ to review the seed data, assessing its quality across different tasks and topics. This allows us to identify deficiencies in the original dataset and strategically generate new data to address these gaps.

Inspired by Constitutional AI \citep{bai2022constitutional}, we define a constitution as a set of principles including clarity, usefulness, challenge, safety, professionalism, and guidance to evaluate instruction-response pairs. For each seed example, $M_t$ conducts multiple reviews of the data against these principles, with each review consisting of a single score from 0 to 10 and a corresponding detailed rationale. This process is formalized as follows:
\begin{equation} \label{eq:review}
    \text{Review}_i = M_t(x_i, y_i; \mathcal{C}), ~ \forall (x_i, y_i) \in D^{\text{seed}}_t
\end{equation}
where $\mathcal{C}$ is the designed constitution, $D^{\text{seed}}_t$ denotes the seed dataset at the current iteration round $t$, and $\text{Review}_i$ encapsulates the evaluation results, including both scores and detailed rationales. we utilize a regular expression $f$ on $\text{Review}_i$ to extract the scores $\mathcal{S}_i = f(\text{Review}_i)$. Next, we calculate the average score $\bar{\mathcal{S}}_i$ from the extracted scores $\mathcal{S}_i$. 

\subsubsection{Reward-Based Generation}
For each seed example, we compare the reward value $\bar{\mathcal{S}}_i$ generated in the previous step to a given threshold $V$. Typically, $V$ is set to 7.0 based on empirical observations. If $\bar{\mathcal{S}}_i$ is below $V$, it indicates that the dataset is poorly distributed across topics or tasks related to this data, potentially lacking high-quality examples. We define the set of low-quality examples as $D_{\text{lq}} = \{ (x_i, y_i) \mid \bar{\mathcal{S}}_i < V \}$.

To address the challenges posed by data distribution deficiencies and enhance model performance, we employ few-shot learning \citep{brown2020language} and Chain-of-Thought (CoT) reasoning \citep{wei2022chain} to prompt the model to generate new instructions based on the original data. Specifically, for every subpar example $(x_i, y_i) \in D_{\text{lq}}$, $K$ new instructions $\{x^{'}_{ij}\}^K_{j=1}=M_t(x_i, y_i;\mathcal{P}^s)$ are created with few-shot prompt $\mathcal{P}^s$. These newly minted instructions are subsequently input into the model $M_t$, yielding fresh responses $y^{'}_{ij}=M_t(x^{'}_{ij})$. The resultant instruction-response pairs $\{(x^{'}_{ij}, y^{'}_{ij})\}^{N, K}_{i,j=1}$ explore the same topics as the original instructions but with higher quality, used for instruction tuning after further processing.

Conversely, if the reward value meets or exceeds $V$, it suggests that this data is effective at instructing the model to accomplish that corresponding task. Formally, we define the set of high-quality examples as $D_{\text{hq}} = \{ (x_i, y_i) \mid \bar{\mathcal{S}}_i \geq V \}$.

To further enhance the dataset, we utilize few-shot learning and COT to prompt the model to generate new, intentionally flawed responses containing misleading information. For each high-quality example $(x_i, y_i) \in D_{\text{hq}}$, we generate $K$ flawed responses $\{y^{\text{fl}}_{ij}\}^K_{j=1} = M_t(x_i, y_i;\mathcal{P}^p)$. These flawed responses and the original ones are jointly sorted based on their reward values, forming preference pairs that can be used for preference learning.

\definecolor{myred}{RGB}{255,0,0}
\definecolor{mygreen}{RGB}{0,128,0}
\definecolor{LightCyan}{rgb}{0.8, 0.9, 1}

\begin{table*}[!t]
\resizebox{\textwidth}{!}{
\begin{tabular}{c|c|c|cccccc}
\toprule
\multirow{2}{*}{\textbf{Backbone}} & \multirow{2}{*}{\textbf{Method}} & \multirow{2}{*}{\textbf{Average}} & \multicolumn{6}{c}{\textbf{HuggingFace Open LLM Leaderboard}} \\ \cline{4-9}
& & & \textbf{ARC} & \textbf{HellaSwag} & \textbf{MMLU} & \textbf{TruthfulQA} & \textbf{GSM8K} & \textbf{Winogrande} \\ \midrule
\multirow{5}{*}{\rotatebox{90}{\textbf{7B}}} & SFT & 64.60 & \underline{51.11} & 78.63 & 68.71 & 55.15 & 60.96 & \underline{73.01} \\ \cline{2-9}
& Self-Instruct & 64.66 \textcolor{myred}{\scriptsize{(+0.06)}} & \textbf{52.39} & 78.34 & 69.19 & 50.30 & 65.20 & 72.53 \\ \cline{2-9}
& SPIN & 68.00 \textcolor{myred}{\scriptsize{(+3.41)}} & 50.43 & \textbf{78.98} & \textbf{69.68} & \underline{55.35} & \underline{81.43} & 72.14 \\ \cline{2-9}
& I-SHEEP & 67.06 \textcolor{myred}{\scriptsize{(+2.46)}} & 51.02 & 78.18 & 68.34 & 53.72 & 78.54 & 72.53 \\ \cline{2-9}
& \cellcolor{LightCyan}\ourmethod & \cellcolor{LightCyan}\textbf{68.24} \textcolor{myred}{\scriptsize{\textbf{(+3.64)}}} & \cellcolor{LightCyan}50.68 & \cellcolor{LightCyan}\underline{78.76} & \cellcolor{LightCyan}\underline{69.31} & \cellcolor{LightCyan}\textbf{55.54} & \cellcolor{LightCyan}\textbf{82.11} & \cellcolor{LightCyan}\textbf{73.01} \\ \midrule \midrule
\multirow{5}{*}{\rotatebox{90}{\textbf{7B-Instruct}}} & SFT & 67.48 & 53.07 & 78.32 & 68.10 & 53.96 & 79.83 & \textbf{71.59} \\ \cline{2-9}
& Self-Instruct & 67.94 \textcolor{myred}{\scriptsize{(+0.47)}} & \underline{54.44} & 79.63 & \textbf{69.94} & 52.67 & 81.05 & 69.93 \\ \cline{2-9}
& SPIN & 68.41 \textcolor{myred}{\scriptsize{(+0.93)}} & 53.07 & 79.72 & \underline{69.80} & 55.27 & 82.03 & 70.56 \\ \cline{2-9}
& I-SHEEP & 68.67 \textcolor{myred}{\scriptsize{(+1.20)}} & 53.16 & \textbf{79.95} & 69.61 & \textbf{57.13} & \underline{82.34} & 69.85 \\ \cline{2-9}
& \cellcolor{LightCyan}\ourmethod & \cellcolor{LightCyan}\textbf{69.22} \textcolor{myred}{\scriptsize{\textbf{(+1.75)}}} & \cellcolor{LightCyan}\textbf{55.89} & \cellcolor{LightCyan}\underline{79.74} & \cellcolor{LightCyan}69.58 & \cellcolor{LightCyan}\underline{55.62} & \cellcolor{LightCyan}\textbf{83.55} & \cellcolor{LightCyan}\underline{70.96} \\ \bottomrule
\end{tabular}}
\caption{\textbf{Performance of multiple self-evolution methods at their optimal iteration rounds across various benchmarks on Qwen2.} SFT represents the initial model obtained through SFT on the seed dataset. \textbf{Bold} values denote the best results achieved, \underline{underlined} values signify the second-best results, \textcolor{myred}{red} values highlight the improvement over the base model. \ourmethod outperforms other baselines in terms of average performance across these benchmarks, often ranking as the top or second-best in most benchmarks.}
\label{tab:main}
\end{table*}

\subsubsection{Data Cleaning and Annotation}

To construct a high-quality training dataset, the generated data is cleaned through two steps: length filtering $\mathbb{L}$ removes overly short or long entries, while similarity filtering $\mathbb{S}$ uses ROUGE-L \citep{lin-2004-rouge} to eliminate data with high similarity to existing instructions or responses. This process mitigates reduced model diversity, as noted by \citet{wu2024progress}. Formally, the process is defined as:
\begin{align}
    (x^{'}_{ij}, y^{'}_{ij})^f &= \mathbb{S}(\mathbb{L}(x^{'}_{ij}, y^{'}_{ij})) \\
    (x_{i}, y_{ij}^\text{fl})^f &= \mathbb{S}(\mathbb{L}(x_{i}, y_{ij}^\text{fl}))
\end{align}
where $(x^{'}_{ij}, y^{'}_{ij})^f$ and $(x_{i}, y_{ij}^\text{fl})^f$ represent $(x^{'}_{ij}, y^{'}_{ij})$ and $(x_{i}, y_{ij}^\text{fl})$ cleaned respectively.

After these computationally inexpensive cleaning steps, we utilize $M_t$ to reward newly generated data via Equation \ref{eq:review}. For instruction tuning, data with a reward $\bar{\mathcal{S}}_i$ greater than $V$ are first collected into the dataset $D_t^S$, which is then merged with the instruction dataset $D^S$. Formally, $D_t^S$ is defined as: $D_t^S = \{(x^{'}_{ij}, y^{'}_{ij})^f \mid \bar{\mathcal{S}}_i \geq V \}$. The instruction dataset $D^S$, initialized from $D^{\text{seed}}_t$, is then updated by merging it with $D^S \gets D^S \cup D_t^S$.

For data intended for preference training, we compare the rewards of each of the two responses. Let $\bar{\mathcal{S}}_i^{a}$ and $\bar{\mathcal{S}}_i^{b}$ denote the reward values of the two responses for the $i$-th instruction. If $\bar{\mathcal{S}}_i^{a} > \bar{\mathcal{S}}_i^{b}$, the first item is used as the preferred response and the second as the dispreferred response; otherwise, the roles are reversed. Formally, the preference pair for the $i$-th instruction is constructed as:
\begin{equation}
    (x_i, y_i^w, y_i^l) = 
    \begin{cases}
        (x_i, y_i^{a}, y_i^{b}) & \text{if } \bar{\mathcal{S}}_i^{a} > \bar{\mathcal{S}}_i^{b}, \\
        (x_i, y_i^{b}, y_i^{a}) & \text{otherwise},
    \end{cases}
\end{equation}
where $x_i$ is the instruction, $y_i^w$ is the preferred response, and $y_i^l$ is the dispreferred response. These preference pairs are collected into a temporary preference dataset $D_t^P= \{(x_i, y_i^w, y_i^l)\}$. The preference dataset $D^P$, initialized as an empty set $\phi$, is then updated by merging it with $D^P \gets D^P \cup D_t^P$.

\subsection{Language Models as Continuous Self-Evolving Data Engineers}

\begin{table*}[htbp]
\centering
\begin{adjustbox}{width=\textwidth}
\begin{tabular}{c|c|ccccccccc}
\toprule
\multirow{2}{*}{\textbf{Model}} & \multirow{2}{*}{\textbf{Average}} & \multirow{2}{*}{\textbf{GSM8K}} & \multirow{2}{*}{\textbf{MATH}} & \multicolumn{5}{c}{\textbf{MGSM\_latin}} & \multirow{2}{*}{\makecell{\textbf{Olympiad} \\ \textbf{Bench}}} & \multirow{2}{*}{\makecell{\textbf{Minerva} \\ \textbf{Math}}} \\
\cline{5-9}
 & & & & \textbf{de} & \textbf{sw} & \textbf{sw} & \textbf{fr} & \textbf{average} & & \\
\midrule
SFT & 40.32 & 60.96 & 41.74 & 52.80 & 1.20 & 60.40 & 58.00 & 47.36 & 11.10 & 44.68 \\ \midrule
Self-Instruct & 31.63 & 65.20 & 11.98 & 29.60 & 6.00 & 42.80 & 34.80 & 28.30 & 8.00 & 44.68 \\ \midrule
SPIN Iter1 & 37.56 & 81.43 & 25.62 & 20.40 & 0.80 & 36.00 & 34.00 & 22.80 & 12.30 & 45.64 \\
SPIN Iter2 & 39.87 & \underline{81.96} & 26.00 & 54.80 & 0.00 & 43.20 & 36.80 & 33.70 & 12.30 & 45.40 \\
SPIN Iter3 & 39.97 & 81.58 & 26.44 & 54.40 & 0.00 & 43.60 & 36.00 & 33.50 & 12.70 & 45.62 \\
SPIN Iter4 & 34.68 & 80.74 & 17.80 & 46.40 & 0.00 & 20.00 & 4.00 & 17.60 & 12.00 & 45.28 \\ \midrule
I-SHEEP Iter1 & 37.86 & 73.09 & 34.50 & 47.20 & 2.00 & 62.40 & 44.00 & 38.90 & 7.10 & 35.70 \\
I-SHEEP Iter2 & 38.05 & 74.37 & 36.38 & 44.00 & 0.80 & 57.60 & 45.60 & 37.00 & 7.10 & 35.38 \\
I-SHEEP Iter3 & 38.57 & 70.96 & 39.80 & 35.20 & 2.80 & 57.20 & 51.60 & 36.70 & 8.10 & 37.28 \\
I-SHEEP Iter4 & 38.96 & 78.54 & 34.40 & 41.60 & 1.60 & 56.40 & 44.40 & 36.00 & 7.40 & 38.46 \\ \midrule
\cellcolor{LightCyan}\ourmethod Iter1 & \cellcolor{LightCyan}43.51 & \cellcolor{LightCyan}67.32 & \cellcolor{LightCyan}41.90 & \cellcolor{LightCyan}65.60 & \cellcolor{LightCyan}2.80 & \cellcolor{LightCyan}70.40 & \cellcolor{LightCyan}\textbf{67.20} & \cellcolor{LightCyan}55.44 & \cellcolor{LightCyan}11.40 & \cellcolor{LightCyan}45.44 \\
\cellcolor{LightCyan}\ourmethod Iter2 & \cellcolor{LightCyan}44.96 & \cellcolor{LightCyan}66.64 & \cellcolor{LightCyan}46.54 & \cellcolor{LightCyan}\underline{66.80} & \cellcolor{LightCyan}3.60 & \cellcolor{LightCyan}\underline{71.60} & \cellcolor{LightCyan}\underline{66.80} & \cellcolor{LightCyan}58.56 & \cellcolor{LightCyan}13.60 & \cellcolor{LightCyan}\underline{45.84} \\
\cellcolor{LightCyan}\ourmethod Iter3 & \cellcolor{LightCyan}\underline{47.54} & \cellcolor{LightCyan}80.14 & \cellcolor{LightCyan}\underline{47.22} & \cellcolor{LightCyan}65.60 & \cellcolor{LightCyan}\textbf{6.80} & \cellcolor{LightCyan}68.00 & \cellcolor{LightCyan}65.20 & \cellcolor{LightCyan}57.52 & \cellcolor{LightCyan}\underline{13.60} & \cellcolor{LightCyan}45.35 \\
\cellcolor{LightCyan}\ourmethod Iter4 & \cellcolor{LightCyan}\textbf{48.83} & \cellcolor{LightCyan}\textbf{82.11} & \cellcolor{LightCyan}\textbf{48.12} & \cellcolor{LightCyan}\textbf{67.60} & \cellcolor{LightCyan}\underline{6.40} & \cellcolor{LightCyan}\textbf{72.00} & \cellcolor{LightCyan}66.40 & \cellcolor{LightCyan}\textbf{59.04} & \cellcolor{LightCyan}\textbf{14.70} & \cellcolor{LightCyan}\textbf{46.10}\\
\bottomrule
\end{tabular}
\end{adjustbox}
\caption{Evolution of mathematical reasoning capabilities in multiple self-evolving algorithms on Qwen2-7B.}
\label{tab:math_performance}
\end{table*}

The core of \ourmethod lies in its ability to continuously refine and evolve language models through iterative data engineering. The end-to-end process is outlined as follows. Initially, the algorithm uses the initial seed data $Seed_0$ and the language model $M$ to perform SFT, resulting in the initial model $M_0$. In each iteration $t$ (from 0 to $N-1$), it employs the method described in Section \ref{sec:data_construction} to generate two datasets $D_t^P$ and $D_t^S$, which are used to update the preference dataset $D^P$ and the supervised dataset $D^S$, respectively. Next, the supervised dataset $D^S$ is then used to fine-tune $M_t$ through SFT, producing the model $M_t^S$. Then, the preference dataset $D^P$ is used for DPO on $M_t^S$, yielding the model $M_t^D$. This model $M_t^D$ is then directly used as the model for the next iteration, denoted $M_{t+1}$. After $N-1$ rounds of iteration, the final model $M_{N}$ obtained is a more powerful language model.

\begin{figure}[!t]
\centering   
\subfigure[Qwen2-7B]{\label{fig:base1}\includegraphics[width=0.46\textwidth]{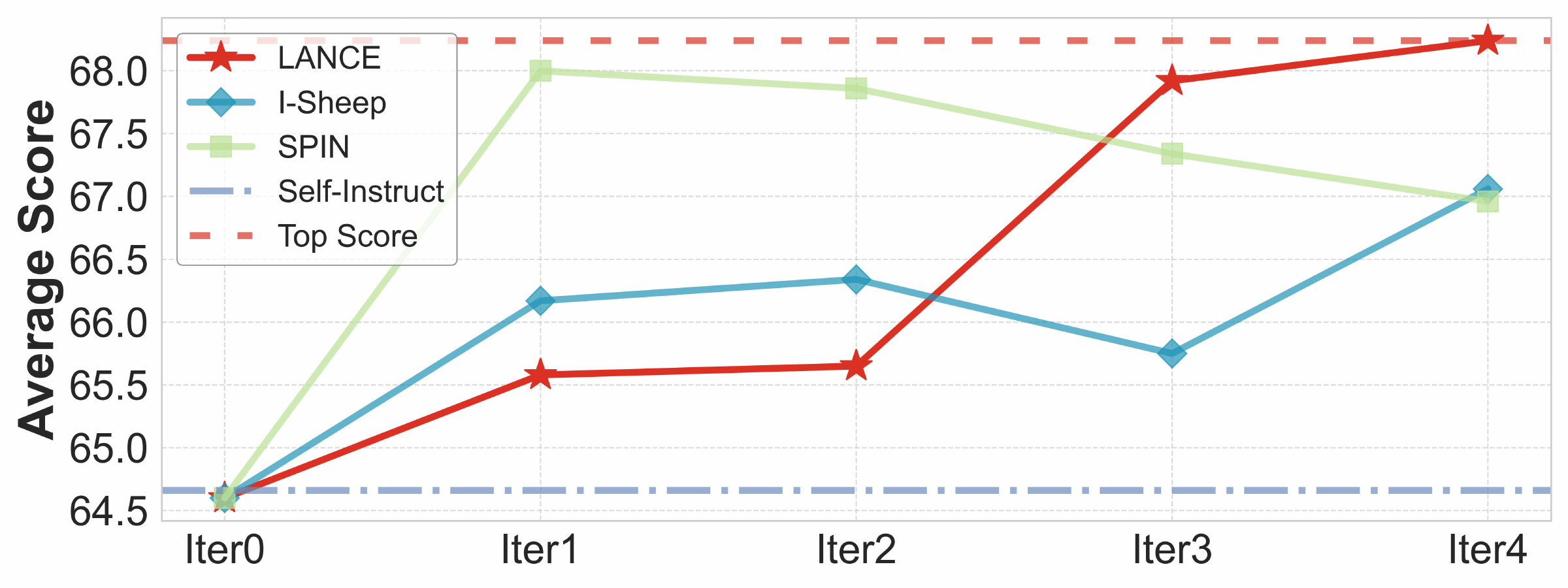}}
\hfill
\subfigure[Qwen2-7B-Instruct]{\label{fig:instruct1}\includegraphics[width=0.46\textwidth]{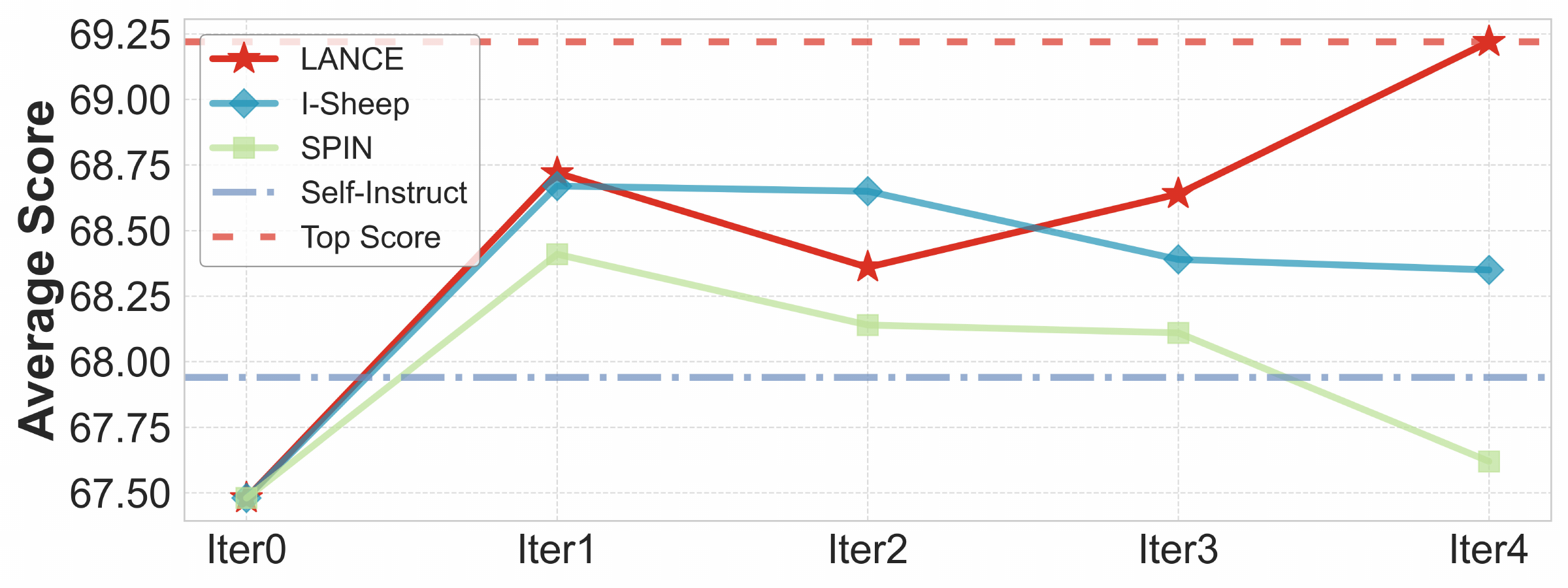}}
\caption{\textbf{Various self-evolution methods show average scores across benchmarks.} The Self-Instruct method, without iterative processes, sampled 50k examples for self-training. "Iter $t$" denotes the $t$-th iteration.}
\label{fig:main}
\end{figure}

\section{Experiments} \label{sec:exp}

\subsection{Experimental Setup}

\textbf{Models} We used Qwen2-7B and Qwen2-7B-Instruct \citep{qwen2} as the backbone models to assess the effectiveness of our training paradigm across different model alignment phases.

\noindent \textbf{Datasets} We construct the seed dataset from two sources: (1) 3,184 examples sampled from UltraChat \citep{ding2023enhancing}, and (2) 5,632 examples from OpenAssistant Conversations \citep{kopf2024openassistant}, which include human-labeled scores. Using Llama3-70B \citep{dubey2024llama}, we generate reward rationales and scores, retaining only those consistent with the human labels.

\noindent \textbf{Baselines} We employ several representative methods as our baselines:
(1) SFT (Supervised Fine-Tuning): The starting point for all self-evolution methods.
(2) Self-Instruct \citep{wang-etal-2023-self-instruct}: 
Enhances instruction-following by generating new instruction data using the LLM itself.
(3) SPIN \citep{chen2024selfplay}: 
A self-evolution approach that leverages self-play fine-tuning to improve the LLM.
(4) I-SHEEP \citep{liang2024sheep}: 
A self-alignment method where the LLM generates and assesses its own training data to achieve self-improvement.

\noindent \textbf{Benchmarks} We assess our model using a comprehensive suite of benchmarks aligned with the Huggingface Open LLM Leaderboard \citep{open-llm-leaderboard}, including HellaSwag \citep{zellers2019hellaswag} and Winogrande \citep{sakaguchi2020winogrande} for commonsense reasoning, MMLU \citep{hendrycks2021measuring} for multi-domain knowledge, TruthfulQA \citep{lin2022truthfulqa} for accuracy, GSM8K \citep{cobbe2021training} for mathematical reasoning, and ARC-Challenge \citep{Clark2018ThinkYH} for scientific reasoning. Additionally, we assess the model's iterative improvements in mathematical reasoning abilities using MATH \citep{hendrycks2021measuring}, OlympiadBench \citep{he2024olympiadbench}, MGSM \citep{shi2023language}, and Minerva Math \citep{lewkowycz2022solving}. 
The specific evaluation settings for these benchmarks are detailed in Appendix \ref{sec:app_par}. Additionally, Appendix \ref{sec:app_hyp} details the hyperparameter settings used during the sampling and training processes.

\subsection{\ourmethod Improves Benchmark Performance}

Table \ref{tab:main} presents the results of \ourmethod and other iterative self-evolution methods across multiple benchmarks on Qwen2, showing the performance at their optimal iteration rounds. Figure \ref{fig:main} illustrates the average performance of these methods across each iteration round. Notably, our approach demonstrates strong performance both when post-training the pre-trained model and when further refining the fully trained model. Appendix \ref{sec:app_iter} shows performance on each benchmark per iteration, and Appendix \ref{sec:app_evo} tracks model performance after each step.

For the average scores, \ourmethod shows an improvement of 3.64 on Qwen2-7B over the initial model (SFT), and 1.75 on Qwen2-7B-Instruct. The most notable improvement is in mathematical abilities. On GSM8K, Qwen2-7B improves by 21.15, while Qwen2-7B-Instruct improves by 3.72. For MMLU, a benchmark covering a wide range of domains and significant challenges, our method also yields improvements: Qwen2-7B improves by 0.60, and Qwen2-7B-Instruct by 1.48. Other abilities either improve or remain largely unchanged.

Regarding iterative self-evolution, our method demonstrates notable improvements over iterations, particularly for the pre-trained model (Qwen2-7B), as shown in the line chart (Figure \ref{fig:main}). For the fully post-trained model (Qwen2-7B-Instruct), while intermediate iterations exhibit some fluctuations, the final iteration achieves a substantial performance gain, surpassing all previous rounds and reaching the highest performance level. This difference likely arises because Qwen2-7B-Instruct, having undergone extensive post-training, requires higher-quality data for further improvement. Consequently, notable gains only emerge after multiple iterations generate sufficiently high-quality data.

In contrast, SPIN shows improvements only in the first round, with performance declining in subsequent rounds. This may be explained by \citet{chen2024selfplay}, who establish that SPIN converges only when the LLM's distribution aligns with the seed data. The limited seed data in our experiments likely restricts SPIN's gains, whereas \ourmethod can generate data beyond the seed distribution, enabling continuous improvement. Further discussion and distribution visualization can be found in Appendix \ref{sec:app_vis}, with case studies in Appendix \ref{sec:app_case}.

I-SHEEP exhibits improvement in the first iteration on Qwen2-7B-Instruct but suffers from performance degradation in later iterations. On Qwen2-7B, I-SHEEP demonstrates an oscillating pattern, with performance rising in iterations 1 and 2, dropping in iteration 3, and recovering in iteration 4. However, even after this recovery, I-SHEEP's final performance remains markedly lower than LANCE's, with a gap of 1.18. Due to computational constraints, we conducted four rounds of iterative experiments with $N=5$. The results show consistent performance improvements in each iteration, indicating potential for further gains with additional iterations and highlighting the promise of self-evolving language models.

\begin{table}[!t]
\centering
{
\begin{tabular}{cccc}
\toprule
\textbf{Model} & \textbf{Full} & \textbf{w/o dpo} & \textbf{w/o sft}  \\ \midrule
SFT & 61.42 & 61.42 & 61.42 \\
\ourmethod Iter1 & 65.58 & 66.89 & \underline{61.57} \\
\ourmethod Iter2 & 65.65 & 67.08 & 61.00 \\
\ourmethod Iter3 & \underline{67.92} & \textbf{67.85} & 59.23 \\
\ourmethod Iter4 & \textbf{68.24} & \underline{67.79} & \textbf{64.08} \\
\bottomrule
\end{tabular}
}
\caption{The changes in average performance of the Qwen2-7B model when certain steps are removed from \ourmethod. SFT serves as the starting point for all iterations. "w/o dpo" excludes DPO-related steps, while "w/o sft" removes SFT-related steps. The results underscore the necessity of a complete pipeline, as the absence of either component leads to slow and unstable improvement.}
\label{tab:ablation}
\end{table}

\subsection{\ourmethod Improves Math Skills}

Table \ref{tab:math_performance} compares Qwen2-7B's performance under various self-evolution algorithms on mathematical reasoning benchmarks. We included GSM8k, a benchmark from the Open LLM Leaderboard that evaluates elementary mathematical abilities, along with MATH, Olympiad Bench, and Minerva Math to assess advanced reasoning capabilities on competition-level and even Olympiad-level mathematical problems. Furthermore, MGSM was used to assess multilingual mathematical proficiency.

On the benchmark of elementary mathematical abilities (as measured by GSM8k), all methods improved the SFT model's performance. \ourmethod also exhibited notable advantages in competition-level mathematical abilities. Specifically, on the MATH benchmark, only \ourmethod achieved a notable improvement, increasing accuracy by 6.38. On the Olympiad Bench, while both \ourmethod and SPIN improved accuracy, SPIN's best result yielded only a modest gain of 1.60, whereas \ourmethod achieved a more substantial improvement of 3.60. Additionally, on the Minerva Math benchmark, both \ourmethod and SPIN enhanced model performance with comparable gains. This discrepancy may stem from the following: improvements in elementary mathematical abilities rely on pattern memorization and optimization of simple problems, while advancements in higher-level mathematical abilities require methods capable of generating data with complex reasoning logic. The accurate generation of data involving complex reasoning may benefit from our approach of requiring the model to engage in thorough deliberation before producing an answer during both the generation and review phases. This deliberate reasoning process enhances the model's understanding of the input, thereby improving the quality and accuracy of the generated data.

In terms of multilingual mathematical abilities (as evaluated by MGSM), despite the training set being exclusively in English, the model also achieved substantial improvements in mathematical proficiency across other Latin-based languages. Notably, only our algorithm demonstrated this cross-lingual transfer capability. This phenomenon may be explained by \ourmethod's ability to generate high-quality mathematical data that genuinely enhances the model's reasoning capabilities, enabling strong generalization across linguistic contexts.

Interestingly, despite the seed dataset being generic and not specifically tailored to mathematical data, \ourmethod still demonstrated a remarkable ability to enhance the model's mathematical reasoning capabilities. This finding highlights the potential of our approach to effectively leverage general-purpose data for improving mathematical problem-solving abilities.

\subsection{Ablation Studies}
\label{sec:abl}

Table~\ref{tab:ablation} illustrates the impact on the Qwen2-7B model's average performance when either SFT-related or DPO-related components are omitted. When the DPO-related components are removed, the model's performance can still improve iteratively in the first three iterations, but at a slower rate compared to the full pipeline, and performance starts to decline after the fourth iteration. This suggests that DPO components are instrumental in accelerating performance gains and also essential for sustaining long-term improvements. When the SFT-related components are removed, the performance improvement becomes highly unstable: after a performance gain in iter1, performance deteriorates in iter2 and iter3, and then improves again in iter4. This instability highlights the critical role of SFT in stabilizing the model's learning process and ensuring consistent progress. These findings collectively emphasize the necessity of a complete and well-integrated data generation and training pipeline. The interplay between SFT and DPO components appears to be synergistic: SFT provides the foundational stability, while DPO drives faster and more sustainable improvements. Table \ref{tab:abl_details} in Appendix \ref{sec:app_abl} provides a detailed breakdown of the model's performance across each test benchmark throughout the iterations. 

\section{Conclusion}

We introduce a novel training paradigm \ourmethod that empowers LLMs to autonomously generate, clean, review, and annotate data with preference information, reducing post-training data construction time and cost. \ourmethod is effective in the post-training of pre-trained models as well as in further post-training of fully trained models, demonstrating continuous improvements in model performance across various tasks and outperforming other self-evolution methods. Notably, even when trained on general-purpose datasets, \ourmethod substantially enhances the model's mathematical capabilities. By ensuring that the generated data aligns with human preferences while reducing the resource requirements for high-quality data creation, we take a step toward addressing a bottleneck in the emergence of superintelligence, laying the groundwork for future systems that can exceed human capabilities.

\section{Limitation}

Our experimental results demonstrate that while \ourmethod substantially enhances the model's mathematical reasoning capabilities, its improvement on knowledge-dependent tasks remains limited. Specifically, in the absence of external supervision signals, the self-evolved data cannot introduce new knowledge beyond the model's existing capacity. This inherent limitation suggests that self-evolution may have restricted potential in augmenting the model's knowledge base, making it more suitable for enhancing weakly knowledge-dependent capabilities. Addressing the enhancement of strongly knowledge-dependent abilities remains an open challenge for future research.

Furthermore, although our self-evolution framework operates with only a small seed dataset and autonomously generates a substantial volume of training data, the process involves multiple iterations of sampling, reviewing, and annotation, which introduces significant computational overhead. This computational cost represents another critical limitation that necessitates further optimization in future work. Addressing this challenge is essential to improve the scalability, efficiency, and practical applicability of the approach, particularly for resource-constrained environments.

\section{Ethic Statement}

In this study, all datasets and models utilized are publicly available and adhere to their respective open-source licenses. The use of these resources is strictly confined to academic research purposes. However, we acknowledge that the misuse of the methodology proposed in this work could potentially lead to the generation of unethical data. This issue is not unique to our approach but is a shared concern across all data generation methods. The potential biases and ethical challenges associated with AI-generated content (AIGC) necessitate ongoing discussion and research within the community to ensure responsible development and deployment of such technologies.

\section*{Acknowledgments}

Thanks everyone!


\bibliography{custom}

\newpage

\appendix

\section*{Appendix}

\section{Data Distribution Visualization}
\label{sec:app_vis}

\begin{figure}[h]
    \centering
    \includegraphics[width=\linewidth]{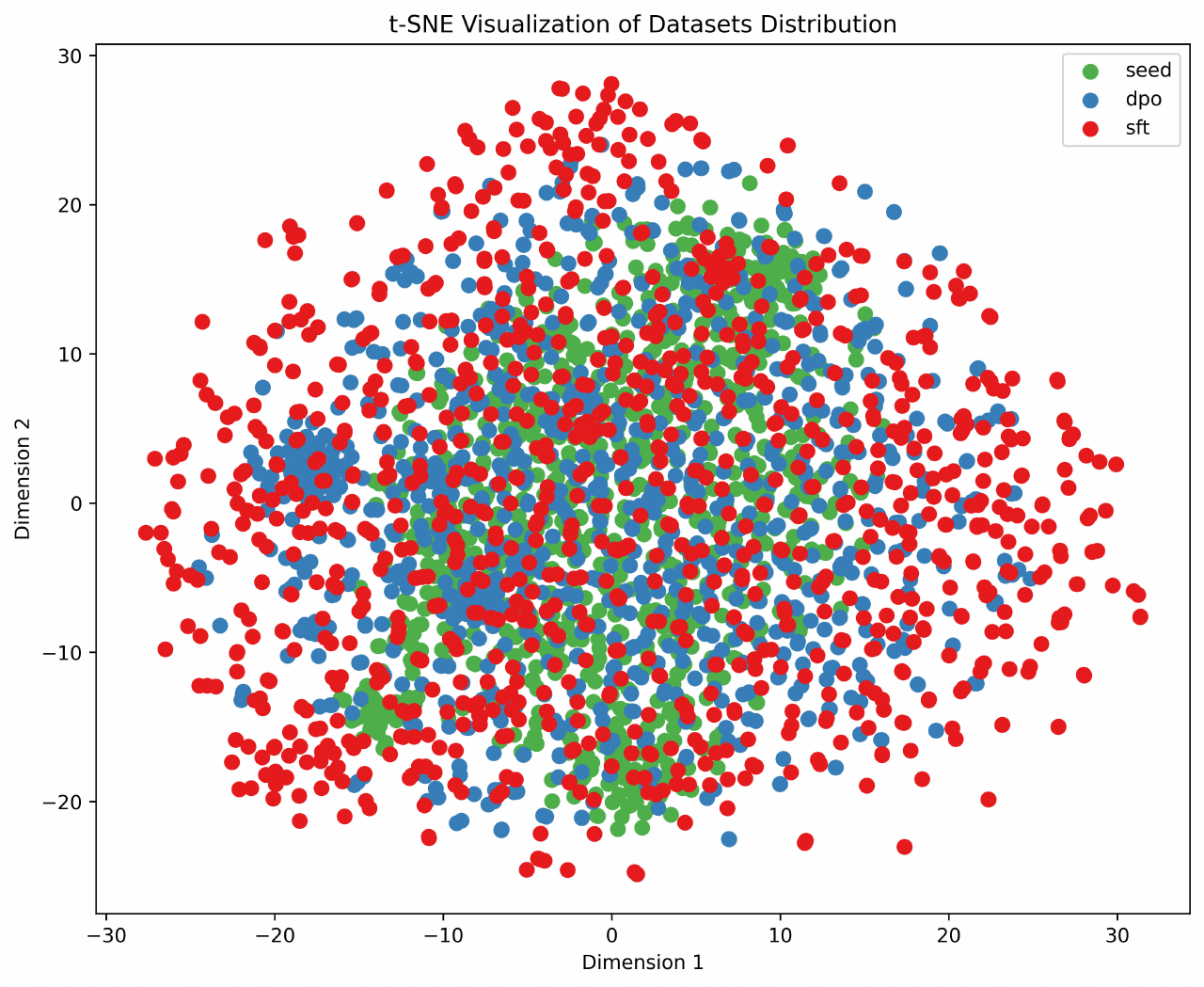}
    \caption{Visualization of the distribution of seed data and synthetic data generated by \ourmethod}
    \label{fig:data_distri}
    \vspace{-1em}
\end{figure}

We sampled 1000 examples each from the seed dataset, the synthetic SFT dataset, and the DPO dataset. Using the stella\_en\_400M\_v5 model \citep{zhang2025jasperstelladistillationsota}, we extracted embeddings for each example and applied t-SNE for dimensionality reduction, visualized in Figure \ref{fig:data_distri}.

The visualization reveals that the synthetic data generated by \ourmethod not only encompasses the distribution range of the original seed data but also explores new regions in the embedding space. This indicates that \ourmethod can produce data that aligns with a broader and more diverse distribution, effectively expanding the original data distribution. Such expansion is particularly valuable for improving model generalization, as it introduces variability that better reflects real-world scenarios.

Notably, the SFT dataset exhibits the widest distribution range among the three datasets. This can be attributed to the generation of new instructions during its construction, which introduces additional diversity and complexity into the data. This suggests that the instruction generation process in \ourmethod plays a critical role in enhancing data diversity and coverage.

Overall, these findings highlight the effectiveness of \ourmethod in generating high-quality synthetic data that not only preserves the characteristics of the original seed data but also extends its boundaries, enabling more robust and generalizable model training.

\section{Evaluation Setting}
\label{sec:app_par}

\begin{table}[!htbp]
\centering
\resizebox{0.48\textwidth}{!}
{
\begin{tabular}{cccc}
\toprule
\textbf{Benchmarks} & \textbf{num shots} & \textbf{version} & \textbf{eval tools} \\ \midrule
ARC-C & 0 & 1.0 & LM Evaluation Harness\footnote{https://github.com/EleutherAI/lm-evaluation-harness} \\
HellaSwag & 0 & 1.0 & LM Evaluation Harness \\
MMLU & 0 & 1.0 & LM Evaluation Harness \\
TruthfulQA & 6 & 2.0 & LM Evaluation Harness \\
Winogrande & 0 & 1.0 & LM Evaluation Harness \\
Minerva Math & 4 & 1.0 & LM Evaluation Harness \\
GSM8k & 4 & 1d7fe4 & OpenCompass\footnote{https://github.com/open-compass/opencompass} \\
MATH & 0 & 393424 & OpenCompass \\
MGSM & 0 & d967bc & OpenCompass \\
Olympiad Bench & 0 & - & Qwen2.5-Math\footnote{https://github.com/QwenLM/Qwen2.5-Math} \\ \bottomrule
\end{tabular}}
\caption{Details of the evaluation settings for all benchmarks in this study.}
\label{tab:app_eval}
\end{table}

\def\thefootnote{1}\footnotetext{\url{https://github.com/EleutherAI/lm-evaluation-harness}}\def\thefootnote{\arabic{footnote}}

\def\thefootnote{2}\footnotetext{\url{https://github.com/open-compass/opencompass}}\def\thefootnote{\arabic{footnote}}

\def\thefootnote{3}\footnotetext{\url{https://github.com/QwenLM/Qwen2.5-Math}}\def\thefootnote{\arabic{footnote}}

Figure \ref{tab:app_eval} provides an overview of the evaluation details for all benchmarks included in this study. The \textbf{'num shots'} column indicates the number of few-shot examples provided during evaluation, which varies across benchmarks, with values such as 0, 4, or 6 depending on the task. The \textbf{'version'} column specifies the version of the evaluation configuration file used, ensuring reproducibility and consistency in the assessment process. Lastly, the \textbf{'eval tools'} column identifies the evaluation frameworks employed, such as LM Evaluation Harness and OpenCompass, along with their respective GitHub repositories for reference. This table encapsulates the key parameters and tools used to ensure a rigorous and standardized evaluation across all benchmarks.

\section{Hyperparameter Settings}
\label{sec:app_hyp}

In this section, we provide detailed descriptions of the hyperparameters used during the sampling and training processes. All experiments were conducted using two NVIDIA RTX A6000 GPUs, each equipped with 48GB VRAM, ensuring efficient data generation and model training. Table \ref{tab:hypersample} outlines the hyperparameters employed during the sampling phase with \ourmethod, which includes key settings such as the \textbf{threshold $V$}, set to 7.0, which acts as the reward cutoff for distinguishing high-quality data from low-quality data. The \textbf{sample size $K$}, set to 4, determines the number of samples generated during both the data generation and review phases. To maintain randomness and control the diversity of outputs, \textbf{Top-p} is set to 0.9, and \textbf{Temperature} is set to 0.7. Additionally, \textbf{Max New Tokens} limits the maximum number of tokens generated during sampling to 512, while \textbf{Min Length} and \textbf{Max Length} define the acceptable token length range for filtering, set to 10 and 4096, respectively.

\begin{table}[!htbp]
\centering
{
\begin{tabular}{lc}
\toprule
\textbf{Parameter} & \textbf{Value} \\ 
\midrule
Threshold $V$ & 7.0 \\
Sample nums $K$ & 4 \\
Top-p & 0.9 \\
Temperature & 0.7 \\
Max new tokens & 512 \\
Min length & 10 \\
Max length & 4096 \\
\bottomrule
\end{tabular}}
\caption{The hyperparameters used during the sampling process with \ourmethod.}
\label{tab:hypersample}
\end{table}

Meanwhile, Table \ref{tab:hypertrain} details the hyperparameters used during the training phase, where the model is fine-tuned on the generated data. For SFT, the learning rate is set to 3e-5, with a batch size of 2 and gradient accumulation steps of 2, trained over 1 epoch. A warmup ratio of 0.01 is applied, and the cutoff length is set to 4096. For DPO, the learning rate is reduced to 5e-6, with a batch size of 1 and gradient accumulation steps of 8, also trained over 1 epoch. The warmup ratio remains at 0.01, and the cutoff length is similarly set to 4096. The $\beta$ parameter, which controls the strength of the preference optimization, is set to 0.2. These configurations ensure a balanced and effective training process, tailored to the specific requirements of each method.

\begin{table*}[!t]
\centering
\resizebox{0.98\textwidth}{!}{
\begin{tabular}{cccccccc}
\toprule
\textbf{Method} & \textbf{Learning Rate} & \textbf{Batch Size} & \textbf{Gradient Accumulation} & \textbf{Epochs} & \textbf{Warmup Ratio} & \textbf{Cutoff Length} & \textbf{$\beta$} \\ \midrule
SFT & 3e-5 & 2 & 2 & 1 & 0.01 & 4096 & - \\
DPO & 5e-6 & 1 & 8 & 1 & 0.01 & 4096 & 0.2 \\ \bottomrule
\end{tabular}}
\caption{The hyperparameters used during the training process with \ourmethod.}
\label{tab:hypertrain}
\end{table*}

\begin{table*}[!ht]
\resizebox{\textwidth}{!}{
\begin{tabular}{c|c|c|cccccc}
\toprule
\multirow{2}{*}{\textbf{Base}} & \multirow{2}{*}{\textbf{Model}} & \multirow{2}{*}{\textbf{Average}} & \multicolumn{6}{c}{\textbf{Benchmarks}} \\ \cline{4-9}
& & & \textbf{ARC} & \textbf{HellaSwag} & \textbf{MMLU} & \textbf{TruthfulQA} & \textbf{GSM8K} & \textbf{Winogrande} \\ \midrule
\multirow{14}{*}{\rotatebox{90}{\textbf{Qwen2-7B}}} & SFT & 64.60 & 51.11 & 78.63 & 68.71 & 55.15 & 60.96 & \underline{73.01} \\ \cline{2-9}
& Self-Instruct 50k & 64.66 \textcolor{myred}{\scriptsize{(+0.06)}} & \textbf{52.39} & 78.34 & 69.19 & 50.30 & 65.20 & 72.53\\ \cline{2-9}
& SPIN Iter1 & 68.00 \textcolor{myred}{\scriptsize{(+3.41)}} & 50.43 & 78.98 & 69.68 & 55.35 & 81.43 & 72.14 \\
& SPIN Iter2 & 67.86 \textcolor{mygreen}{\scriptsize{\textbf{(-0.14)}}} & 49.83 & 79.13 & \textbf{69.73} & 55.38 & \underline{81.96} & 71.11 \\
& SPIN Iter3 & 67.34 \textcolor{mygreen}{\scriptsize{\textbf{(-0.52)}}} & 48.12 & \underline{79.31} & \underline{69.71} & 54.99 & 81.58 & 70.32 \\
& SPIN Iter4 & 66.96 \textcolor{mygreen}{\scriptsize{\textbf{(-0.38)}}} & 46.16 & \textbf{79.38} & 69.66 & 55.66 & 80.74 & 70.17 \\ \cline{2-9}
& I-SHEEP Iter1 & 66.17 \textcolor{myred}{\scriptsize{(+1.57)}} & 50.85 & 78.32 & 68.45 & 53.60 & 73.09 & 72.69 \\
& I-SHEEP Iter2 & 66.34 \textcolor{myred}{\scriptsize{(+0.17)}} & \underline{51.45} & 78.27 & 68.49 & 53.76 & 74.37 & 71.67 \\
& I-SHEEP Iter3 & 65.75 \textcolor{mygreen}{\scriptsize{\textbf{(-0.58)}}} & 51.11 & 78.09 & 68.29 & 53.85 & 70.96 & 72.22 \\
& I-SHEEP Iter4 & 67.06 \textcolor{myred}{\scriptsize{(+1.30)}} & 51.02 & 78.18 & 68.34 & 53.72 & 78.54 & 72.53 \\ \cline{2-9}
& \cellcolor{LightCyan}\ourmethod Iter1 & \cellcolor{LightCyan}65.58 \textcolor{myred}{\scriptsize{\textbf{(+0.99)}}} & \cellcolor{LightCyan}50.85 & \cellcolor{LightCyan}78.45 & \cellcolor{LightCyan}68.96 & \cellcolor{LightCyan}55.53 & \cellcolor{LightCyan}67.32 & \cellcolor{LightCyan}72.38 \\
& \cellcolor{LightCyan}\ourmethod Iter2 & \cellcolor{LightCyan}65.65 \textcolor{myred}{\scriptsize{\textbf{(+0.07)}}} & \cellcolor{LightCyan}50.51 & \cellcolor{LightCyan}78.94 & \cellcolor{LightCyan}69.41 & \cellcolor{LightCyan}\textbf{55.97} & \cellcolor{LightCyan}66.64 & \cellcolor{LightCyan}72.45 \\
& \cellcolor{LightCyan}\ourmethod Iter3 & \cellcolor{LightCyan}67.92 \textcolor{myred}{\scriptsize{\textbf{(+2.26)}}} & \cellcolor{LightCyan}50.77 & \cellcolor{LightCyan}79.12 & \cellcolor{LightCyan}69.24 & \cellcolor{LightCyan}\underline{55.78} & \cellcolor{LightCyan}80.14 & \cellcolor{LightCyan}72.45 \\
& \cellcolor{LightCyan}\ourmethod Iter4 & \cellcolor{LightCyan}\textbf{68.24} \textcolor{myred}{\scriptsize{\textbf{(+0.32)}}} & \cellcolor{LightCyan}50.68 & \cellcolor{LightCyan}78.76 & \cellcolor{LightCyan}69.31 & \cellcolor{LightCyan}55.54 & \cellcolor{LightCyan}\textbf{82.11} & \cellcolor{LightCyan}\textbf{73.01} \\ \midrule \midrule
\multirow{14}{*}{\rotatebox{90}{\textbf{Qwen2-7B-Instruct}}} & SFT & 67.48 & 53.07 & 78.32 & 68.10 & 53.96 & 79.83 & \textbf{71.59} \\ \cline{2-9}
& Self-Instruct 50k & 67.94 \textcolor{myred}{\scriptsize{(+0.47)}} & 54.44 & 79.63 & \textbf{69.94} & 52.67 & 81.05 & 69.93 \\ \cline{2-9}
& SPIN Iter1 & 68.41 \textcolor{myred}{\scriptsize{(+0.93)}} & 53.07 & 79.72 & \underline{69.80} & 55.27 & 82.03 & 70.56 \\
& SPIN Iter2 & 68.14 \textcolor{mygreen}{\scriptsize{\textbf{(-0.27)}}} & 51.79 & 79.66 & 69.65 & 56.46 & 81.58 & 69.69 \\
& SPIN Iter3 & 68.11 \textcolor{mygreen}{\scriptsize{\textbf{(-0.03)}}} & 51.19 & \underline{79.99} & 69.49 & 56.42 & 82.11 & 69.46 \\
& SPIN Iter4 & 67.62 \textcolor{mygreen}{\scriptsize{\textbf{(-0.49)}}} & 50.77 & \textbf{80.04} & 69.59 & 56.07 & 81.05 & 68.19 \\ \cline{2-9}
& I-SHEEP Iter1 & 68.67 \textcolor{myred}{\scriptsize{(+1.20)}} & 53.16 & 79.95 & 69.61 & \textbf{57.13} & 82.34 & 69.85 \\
& I-SHEEP Iter2 & 68.65 \textcolor{mygreen}{\scriptsize{\textbf{(-0.02)}}} & 54.10 & 79.92 & 69.56 & \underline{56.53} & 81.96 & 69.85 \\
& I-SHEEP Iter3 & 68.39 \textcolor{mygreen}{\scriptsize{\textbf{(-0.27)}}} & 53.33 & 79.96 & 69.38 & 55.80 & 82.56 & 69.30 \\
& I-SHEEP Iter4 & 68.35 \textcolor{mygreen}{\scriptsize{\textbf{(-0.04)}}} & 53.50 & 79.61 & 69.37 & 55.43 & \underline{82.71} & 69.46 \\ \cline{2-9}
& \cellcolor{LightCyan}\ourmethod Iter1 & \cellcolor{LightCyan}68.72 \textcolor{myred}{\scriptsize{\textbf{(+1.24)}}} & \cellcolor{LightCyan}\underline{55.38} & \cellcolor{LightCyan}79.53 & \cellcolor{LightCyan}69.34 & \cellcolor{LightCyan}55.85 & \cellcolor{LightCyan}81.73 & \cellcolor{LightCyan}70.48 \\
& \cellcolor{LightCyan}\ourmethod Iter2 & \cellcolor{LightCyan}68.36 \textcolor{mygreen}{\scriptsize{\textbf{(-0.36)}}} & \cellcolor{LightCyan}55.12 & \cellcolor{LightCyan}79.76 & \cellcolor{LightCyan}69.55 & \cellcolor{LightCyan}55.98 & \cellcolor{LightCyan}80.14 & \cellcolor{LightCyan}69.61 \\
& \cellcolor{LightCyan}\ourmethod Iter3 & \cellcolor{LightCyan}68.64 \textcolor{myred}{\scriptsize{\textbf{(+0.28)}}} & \cellcolor{LightCyan}55.03 & \cellcolor{LightCyan}79.92 & \cellcolor{LightCyan}69.56 & \cellcolor{LightCyan}55.69 & \cellcolor{LightCyan}80.59 & \cellcolor{LightCyan}\underline{71.03} \\
& \cellcolor{LightCyan}\ourmethod Iter4 & \cellcolor{LightCyan}\textbf{69.22} \textcolor{myred}{\scriptsize{\textbf{(+0.59)}}} & \cellcolor{LightCyan}\textbf{55.89} & \cellcolor{LightCyan}79.74 & \cellcolor{LightCyan}69.58 & \cellcolor{LightCyan}55.62 & \cellcolor{LightCyan}\textbf{83.55} & \cellcolor{LightCyan}70.96 \\ \bottomrule
\end{tabular}}
\caption{\textbf{Experimental results of multiple self-evolution methods across various benchmarks.} \textcolor{myred}{Red} / \textcolor{mygreen}{Green} values indicate improvements / decreases compared to the previous iteration. \ourmethod consistently shows performance gains across iterations, outperforming other baselines.}
\label{tab:main_details}
\end{table*}

\section{Iteration details}
\label{sec:app_iter}

Table \ref{tab:main_details} presents the specific scores of \ourmethod and baseline methods during the iterative process across various benchmarks, where \textcolor{myred}{red} / \textcolor{mygreen}{\textbf{green}} values indicate improvements/declines compared to the previous iteration, respectively. For the pre-trained model Qwen2-7B, only \ourmethod achieved continuous performance improvement (all change values are marked in \textcolor{myred}{red}). Although the I-SHEEP method showed an upward trend in the first two iterations, it experienced a decline in the third iteration. Despite rebounding to the peak in the fourth iteration, its overall performance still lags significantly behind \ourmethod. For the fully post-trained Qwen2-7B-Instruct model, only \ourmethod demonstrated sustained improvement over multiple iterations, while other methods showed progress only in the first iteration. Notably, although \ourmethod exhibited a temporary performance fluctuation in the second iteration, it rebounded in the third iteration and reached the performance peak in the final iteration. These comparative results fully validate the significant advantages and potential of \ourmethod in continuously optimizing model performance.

\begin{table*}[!ht]
\resizebox{\textwidth}{!}{
\begin{tabular}{c|c|c|cccccc}
\toprule
\multirow{2}{*}{\textbf{Base}} & \multirow{2}{*}{\textbf{Model}} & \multirow{2}{*}{\textbf{Average}} & \multicolumn{6}{c}{\textbf{Benchmarks}} \\ \cline{4-9}
                                &                                   &                                   & \textbf{ARC} & \textbf{HellaSwag} & \textbf{MMLU} & \textbf{TruthfulQA} & \textbf{GSM8K} & \textbf{Winogrande}  \\ \midrule
\multirow{9}{*}{\rotatebox{90}{\textbf{Qwen2-7B}}} & SFT           & 64.60                             & 53.07                         & 78.32                               & 68.10                          & 53.96                                & 79.83                                & 71.59                                                  \\ \cline{2-9}
                                
                                &  \ourmethod SFT Iter1                      & 66.89                             & 54.98                         & 79.42                               & 69.44                          & 55.09                                & 81.58                                & 69.85                         \\
                                & \ourmethod DPO Iter1                      & 65.58                             & 55.38                         & 79.53                               & 69.34                          & 55.85                                & 81.73                                & 70.48                        \\
                                &  \ourmethod SFT Iter2                      & 67.94                           & 54.01                         & 79.64                               & 69.61                          & 55.00                                & 81.12                                & 69.46                         \\
                                &  \ourmethod DPO Iter2                      & 65.65                            & 55.12                         & 79.76                               & 69.55                          & 55.98                                & 80.14                                & 69.61                         \\
                                &  \ourmethod SFT Iter3                      & 68.09                             & 54.01                         & 79.52                               & 69.71                          & 54.76                                & 81.27                                & 70.17                         \\
                                &  \ourmethod DPO Iter3                      & 67.92                            & 55.03                         & 79.92                               & 69.56                          & 55.69                                & 80.59                                & 71.03                      \\
                                & \ourmethod SFT Iter4                      & 68.20                             & 54.61                         & 79.49                               & 69.53                          & 55.41                                & 83.17                                & 71.11                        \\
                                & \ourmethod DPO Iter4                      & 68.24                            & 55.89                         & 79.74                               & 69.58                          & 55.62                                & 83.55                                & 70.96 
                      \\\midrule
\midrule
\multirow{9}{*}{\rotatebox{90}{\textbf{Qwen2-7B-Instruct}}} & SFT                    & 67.48                             & 51.11                         & 78.63                               & 68.71                          & 55.15                                & 60.96                                & 73.01                                \\ \cline{2-9}
                                &  \ourmethod SFT Iter1                      & 68.39                             & 50.85                         & 78.36                               & 68.90                          & 54.80                                & 75.74                                & 72.69   \\
                                &  \ourmethod DPO Iter1                      & 68.72                            & 50.85                         & 78.45                               & 68.96                          & 55.53                                & 67.32                                & 72.38                                                   \\
                                &  \ourmethod SFT Iter2                      & 68.14                          & 50.85                         & 78.77                               & 69.01                          & 55.07                                & 81.35                                & 72.61                                                    \\
                                &  \ourmethod DPO Iter2                      & 68.36                        & 50.51                         & 78.94                               & 69.41                          & 55.97                                & 66.64                                & 72.45                        \\
                                &  \ourmethod SFT Iter3                      & 68.24                             & 50.43                         & 78.78                               & 69.42                          & 55.23                                & 81.80                                & 72.85                          \\
                                &  \ourmethod DPO Iter3                      & 68.64                             & 50.77                         & 79.12                               & 69.24                          & 55.78                                & 80.14                                & 72.45                        \\
                                &  \ourmethod SFT Iter4                      & 68.89                           & 50.94                         & 78.64                               & 69.41                          & 55.41                                & 81.96                                & 72.85                         \\
                                &  \ourmethod DPO Iter4                      & 69.22                           & 50.68                         & 78.76                               & 69.31                          & 55.54                                & 82.11                                & 73.01   
                       \\ \bottomrule
\end{tabular}}
\caption{\textbf{Evolution of Model Performance During the Implementation of \ourmethod.} \ourmethod SFT Iter $t$ Denotes the Model Fine-tuned with SFT Data After the $t$-th Iteration, and \ourmethod DPO Iter $t$ Represents the Model Fine-tuned with DPO Data After the $t$-th Iteration.}
\label{tab:step_details}
\end{table*}

\begin{table*}[!ht]
\resizebox{\textwidth}{!}{
\begin{tabular}{c|c|cccccc}
\toprule
\multirow{2}{*}{\textbf{Model}} & \multirow{2}{*}{\textbf{Average}} & \multicolumn{6}{c}{\textbf{Benchmarks}} \\ \cline{3-8}
                          &                                   & \textbf{ARC} & \textbf{HellaSwag} & \textbf{MMLU} & \textbf{TruthfulQA} & \textbf{GSM8K} & \textbf{Winogrande}  \\ \midrule
SFT                             & 64.60                             & 51.11                         & 78.63                               & 68.71                          & 55.15     & 60.96                           & \underline{73.01}                                                           \\ \midrule
LANCE Iter1 w/o dpo             & 66.89                             & 50.85                         & 78.36                               & 68.90                          & 54.80       & 75.74                         & 72.69                                                           \\
LANCE Iter2 w/o dpo             & 67.08                             & 48.63                         & 78.21                               & 69.11                          & 53.87        & 80.97                         & 71.67                                                          \\
LANCE Iter3 w/o dpo             & 67.85                             & 50.51                         & 78.65                               & 69.03                          & 54.78         & \underline{81.80}                         & 72.30                                                         \\
LANCE Iter4 w/o dpo             & 67.79                             & 50.85                         & 78.63                               & 69.06                          & 54.71         & 81.43                        & 72.06                                                          \\ \midrule
LANCE Iter1 w/o sft             & 61.57                             & \textbf{53.05}                         & 78.92                               & 68.98                          & 56.28                    & 39.58                 & 72.61                                                      \\
LANCE Iter2 w/o sft             & 61.00                             & 51.54                         & 78.84                               & 68.89                          & \textbf{56.73}                          & 37.68      & 72.30                                                           \\
LANCE Iter3 w/o sft             & 59.23                             & \underline{51.88}                         & 78.87                               & 68.98                          & \underline{56.56}                 & 26.46                & 72.61                                                          \\
LANCE Iter4 w/o sft             & 64.08                             & 51.79                         & \underline{78.94}                               & 69.06                          & 56.22                       & 56.25         & 72.22                                                           \\ \midrule
LANCE Iter1                     & 65.58                             & 50.85                         & 78.45                               & 68.96                          & 55.53                                & 72.38                                & 67.32                           \\
LANCE Iter2                     & 65.65                             & 50.51                         & 78.94                               & \textbf{69.41}                          & 55.97                   & 66.64             & 72.45                                                           \\
LANCE Iter3                     & \underline{67.92}                             & 50.77                         & \textbf{79.12}                               & 69.24                          & 55.78                      & 80.14          & 72.45                                                           \\
LANCE Iter4                     & \textbf{68.24}                             & 50.68                         & 78.76                               & \underline{69.31}                          & 55.54                 & \textbf{82.11}                & \textbf{73.01}                                       \\                   \bottomrule
\end{tabular}}
\caption{The performance of \ourmethod without SFT-related and DPO-related components on each evaluation benchmark at every iteration step.}
\label{tab:abl_details}
\end{table*}

\section{Model Evolution Steps}
\label{sec:app_evo}

Table \ref{tab:step_details} illustrates the impact of each step on model performance during the implementation of \ourmethod. For Qwen2-7B, although the DPO phase in the first three iterations resulted in performance degradation, the results in Table \ref{tab:step_details} without DPO-related components reveal that the absence of these components leads to only slow and unstable iterative improvements. This underscores the critical role of DPO-related components in enhancing the model's ability to generate high-quality data. For Qwen2-7B-Instruct, performance declined in the second and third iterations but rebounded notably in the fourth iteration, achieving substantial improvement. These findings collectively emphasize the importance of a complete iterative training pipeline for achieving robust and consistent performance gains.

\begin{figure*}[!ht]
    \centering
    \includegraphics[width=\linewidth]{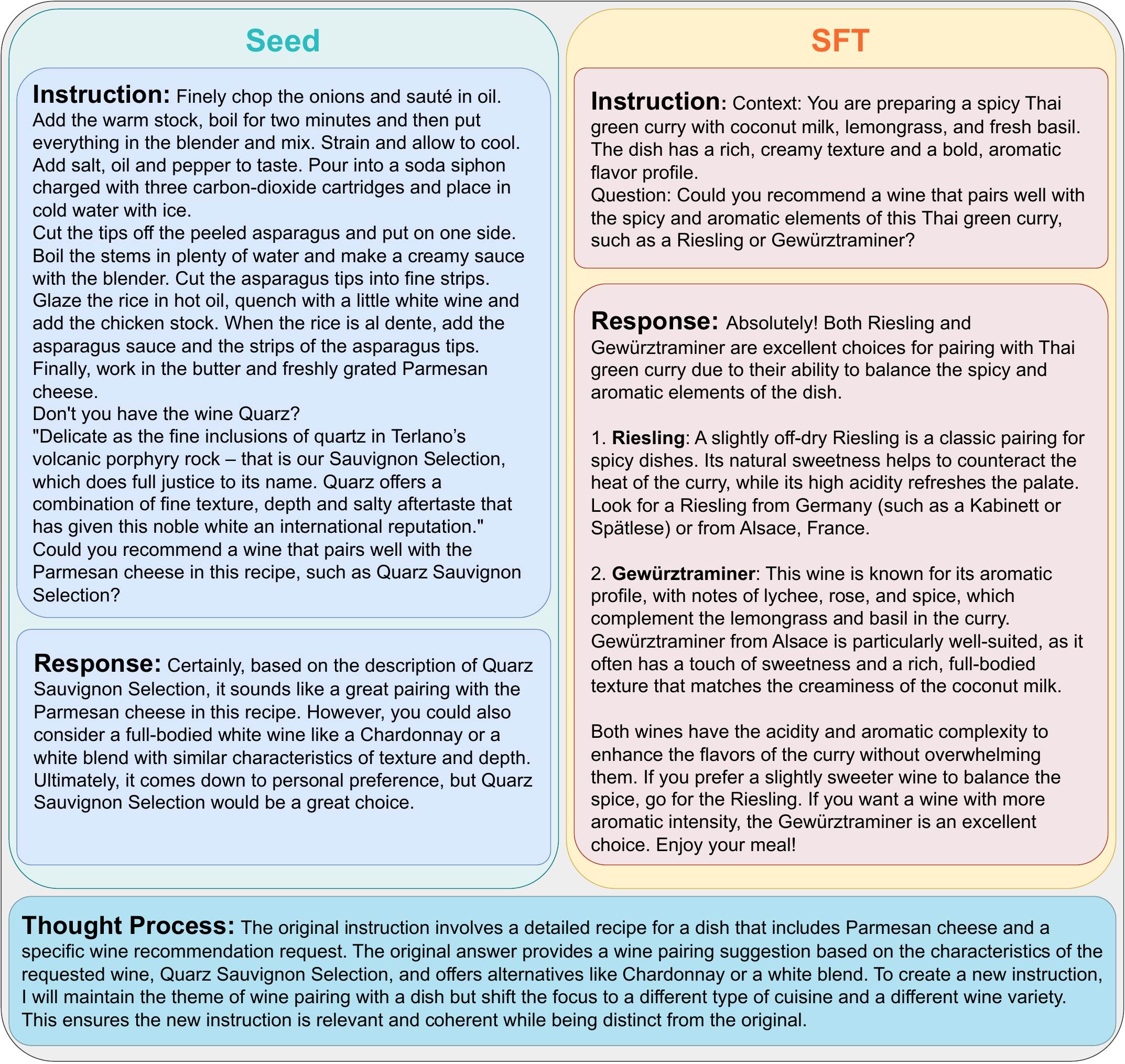}
    \caption{An example of SFT data generation based on seed data using \ourmethod.}
    \label{fig:sft_case}
    \vspace{-1em}
\end{figure*}

\begin{figure*}[!ht]
    \centering
    \includegraphics[width=\linewidth]{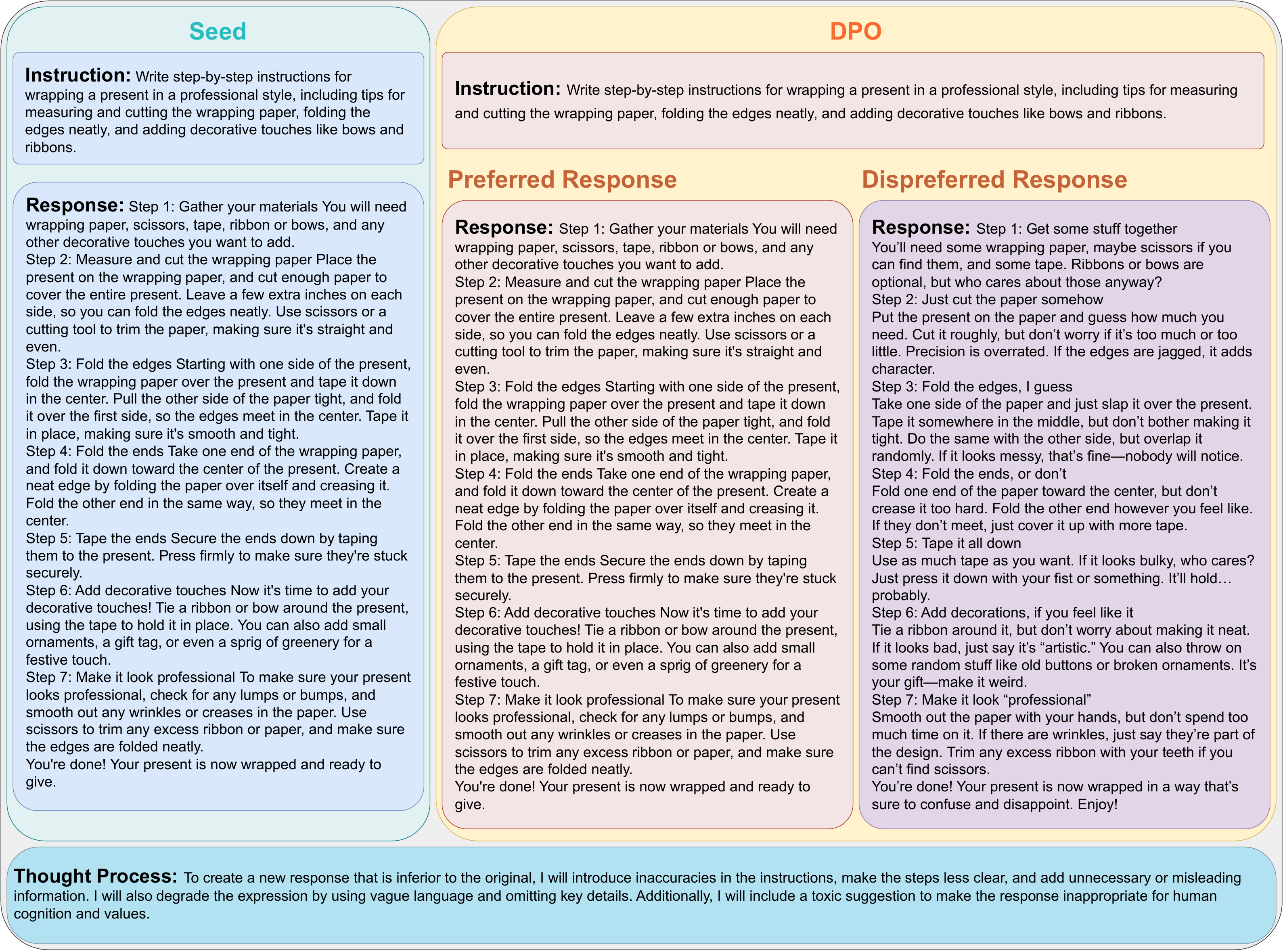}
    \caption{An example of DPO data generation based on seed data using \ourmethod. The Preferred Response and Dispreferred Response represent $y^w$ and $y^l$ in Equation \ref{eq:dpo}, respectively.}
    \label{fig:dpo_case}
    \vspace{-1em}
\end{figure*}

\section{Ablation Details}
\label{sec:app_abl}
Table \ref{tab:abl_details} provides a detailed breakdown of our ablation experiments, showing the performance of \ourmethod with specific components removed across each iteration step and evaluation benchmark. Beyond the findings discussed in Section \ref{sec:abl}, where we highlighted that an incomplete pipeline leads to unstable performance improvements or even declines, we observe that removing SFT-related components significantly degrades the model's performance on GSM8k. This is likely because our SFT data construction process generates novel instructions, which enhances the model's mathematical and logical reasoning capabilities. In contrast, the DPO data construction process does not produce new instructions, which may explain its limited impact in this regard.

\section{Case Study}
\label{sec:app_case}

Figure \ref{fig:sft_case} illustrates the comparison between the SFT data generated from reference seed data and the original seed data. The original seed data exhibits a disorganized structure in its instructions, with the cooking steps and wine recommendations lacking logical coherence. Specifically, the descriptions of wine choices are disconnected from the cooking procedures, creating an abrupt and disjointed flow. In contrast, the SFT data demonstrates a more structured and contextually coherent approach. The generated instruction is carefully designed to align the wine recommendation with the specific flavor profile and culinary context of the dish, ensuring a seamless integration of the two elements.

The original seed data presents a recipe for a dish involving Parmesan cheese and abruptly transitions to a wine recommendation request, which, while relevant, feels disconnected from the preceding cooking steps. The response, though informative, does not explicitly tie the wine suggestions back to the dish's preparation or flavor profile. On the other hand, the SFT data introduces a clear context—preparing a spicy Thai green curry—and explicitly links the wine recommendation to the dish's aromatic and spicy characteristics. This creates a more natural and engaging flow, as the wine pairing is presented as an integral part of the culinary experience rather than an afterthought.

The thought process behind the SFT data generation played a critical role in achieving this improvement. By analyzing the original seed data, we identified its key limitations, such as the lack of logical coherence and contextual integration between the cooking steps and wine recommendations. The thought process guided the creation of a new instruction that maintained the theme of wine pairing but shifted the focus to a different cuisine and wine variety, ensuring relevance and distinctiveness. This deliberate approach allowed us to craft a more cohesive and engaging instruction, where the wine recommendation is tightly coupled with the dish's flavor profile and preparation context.

Furthermore, the SFT data enhances the instructional quality by providing a detailed rationale for the wine recommendations. It not only suggests Riesling and Gewürztraminer but also explains how their specific attributes—such as the sweetness and acidity of Riesling or the aromatic intensity of Gewürztraminer—complement the dish's flavors. This level of detail enriches the user's understanding and provides actionable insights, making the generated data more practical and informative compared to the original seed data. The thought process ensured that these explanations were not only accurate but also tailored to the specific context of the dish, further enhancing the overall quality of the generated content.

In summary, the SFT data improves upon the original seed data by introducing a clear contextual framework, ensuring logical coherence between the cooking instructions and wine recommendations, and providing detailed explanations that enhance the user's engagement and understanding. The thought process was instrumental in achieving these improvements, as it guided the generation of contextually relevant, structured, and user-centric content. This structured and context-aware approach reflects the effectiveness of the method used to generate the SFT data, highlighting its potential for creating high-quality instructional content.

Figure \ref{fig:dpo_case} presents an example of generating a suboptimal response from high-quality seed data to construct DPO data. Notably, this suboptimal response is also generated by \ourmethod, which intentionally introduces imperfections to create a dispreferred response for contrastive learning. This case highlights the process of constructing preference pairs, where the original high-quality response serves as the preferred response, and the generated suboptimal response acts as the dispreferred counterpart. The thought process behind this generation plays a critical role in ensuring the dispreferred response effectively contrasts with the preferred one, providing a clear learning signal for the model.

The original instruction requests step-by-step guidance for wrapping a present in a professional style, including tips for measuring, cutting, folding, and adding decorative touches. The \textbf{preferred response} provides a detailed, structured, and professional set of instructions. It begins with gathering materials, followed by precise steps for measuring and cutting the wrapping paper, folding edges neatly, and securing the ends with tape. The response also emphasizes adding decorative touches like bows and ribbons, and concludes with tips for ensuring a polished, professional appearance. This high-quality reply is clear, actionable, and aligns perfectly with the instruction's intent, making it an ideal candidate for the preferred response in DPO training.

In contrast, the \textbf{dispreferred response} is intentionally designed to deviate from the original's quality and professionalism. The thought process guiding its creation focused on introducing inaccuracies, reducing clarity, and adding unnecessary or misleading information. For example, the response suggests cutting the paper roughly, folding edges haphazardly, and using excessive tape without concern for neatness. It also trivializes the importance of decorative touches, recommending random or unconventional additions like old buttons or broken ornaments. Furthermore, the response dismisses the need for precision or professionalism, suggesting that wrinkles and jagged edges add character or artistic flair. These deliberate deviations were introduced to degrade the quality of the response, making it less useful and less aligned with the instruction's intent.

In summary, this case demonstrates the intentional creation of a suboptimal response from high-quality seed data to form a preference pair for DPO training. The preferred response exemplifies clarity, precision, and professionalism, while the dispreferred response showcases carelessness and a lack of detail. Guided by the thought process, the dispreferred response introduces inaccuracies, vague language, and unnecessary information, creating a clear contrast with the preferred response. Together, these responses provide a valuable contrastive pair, enabling the model to learn and prioritize high-quality outputs, ultimately improving its ability to generate user-aligned content.

\end{document}